\documentclass[runningheads]{llncs}
\usepackage{graphicx,verbatim}
\usepackage{algorithm}
\usepackage{algorithmic}

\usepackage[utf8]{inputenc} 
\usepackage[T1]{fontenc}    
\usepackage{url}            
\usepackage{booktabs}       
\usepackage{amsfonts}       
\usepackage{nicefrac}       
\usepackage{microtype}      
\usepackage{lipsum}
\usepackage{fancyhdr}       
\graphicspath{{media/}}     

\usepackage{textcomp}
\usepackage{siunitx} 
\usepackage{array}   
\usepackage{multirow}
\usepackage{tabularx}
\usepackage{amsmath}
\usepackage{pifont}

\begin{document}
\title{SRMA-Mamba: Spatial Reverse Mamba Attention Network for Pathological Liver Segmentation in MRI Volumes}

\author{
Jun Zeng\inst{1} 
\and Quoc-Huy Trinh\inst{3}
\and Deepak Ranjan Nayak\inst{4}
\and Nikhil Kumar Tomar\inst{5}
\and Ulas Bagci\inst{5}
\and Debesh Jha\inst{2} \thanks{Corresponding author.}
}


\authorrunning{J. Zeng et al.}

\institute{
Chongqing University of Posts and Telecommunications, China
\and
Biomedical Perception \& Intelligence Lab, University of South Dakota, USA
\and
Aalto University, Finland
\and
Malaviya National Institute of Technology Jaipur, India
\and
Machine \& Hybrid Intelligence Lab, Northwestern University, USA
\email{debesh.jha@usd.edu}
}

\maketitle              
\begin{abstract}
Liver cirrhosis plays a critical role in the prognosis of chronic liver disease. Early detection and timely intervention are essential for reducing mortality rates. However, the intricate anatomical architecture and diverse pathological changes of liver tissue complicate the accurate detection and characterization of pathological liver structures in clinical settings. Existing methods underutilize spatial anatomical details in volumetric MRI data, thereby hindering their clinical effectiveness and explainability. To address this challenge, we introduce a novel Mamba-based network, SRMA-Mamba, designed to model the spatial relationships within complex anatomical structures of MRI volumes. By integrating the Spatial Anatomy-Based Mamba module (SABMamba), SRMA-Mamba performs selective Mamba scans within pathological liver tissues and combines anatomical information from the sagittal, coronal, and axial planes to construct a global spatial context representation, enabling efficient volumetric segmentation of pathological liver structures. Furthermore, we introduce the Spatial Reverse Mamba Attention module (SRMA), designed to progressively refine boundary details in the segmentation map, utilizing both the coarse segmentation map and hierarchical encoding features. Extensive experiments demonstrate that SRMA-Mamba surpasses state-of-the-art methods, delivering exceptional performance in 3D pathological liver segmentation. The source code is available at
\url{https://github.com/JunZengz/SRMA-Mamba}.


\end{abstract}
\section{Introduction}
\label{sec:intro}
Cirrhosis constitutes a significant global health burden, ranking as the 11th leading cause of mortality worldwide in 2019~\cite{huang2023global}. As the end-stage manifestation of progressive liver fibrosis, its precise assessment is paramount for determining clinical prognosis and intervention strategies. While liver biopsy remains the diagnostic gold standard~\cite{chowdhury2023liver}, its clinical utility is frequently hampered by its invasive nature and inherent risks, such as post-procedural hemorrhage. Consequently, Magnetic Resonance Imaging (MRI) has emerged as a cornerstone of non-invasive diagnostics~\cite{tonan2010chronic}, leveraging superior soft-tissue contrast to characterize cirrhotic architecture.  However, manual delineation of pathological liver structures from high-dimensional MRI volumes is labor-intensive and susceptible to significant inter-observer variability, underscoring an urgent need for high-precision automated segmentation algorithms.

Recently, state-space models (SSMs)~\cite{gu2021efficiently,smith2022simplified,gu2023mamba} have demonstrated exceptional performance in capturing long-range dependencies. Mamba~\cite{gu2023mamba} employs a selection mechanism and a hardware-aware algorithm to model global dependencies, achieving state-of-the-art performance in linear-time sequence modeling with linear complexity. To extend Mamba into the vision domain, VMamba~\cite{liu2024vmamba} performs four distinct traversal path scans to achieve a 2D global receptive field. Based on this foundation, Mamba-based methods~\cite{ruan2024vm,wang2024mambaunet,xing2024segmamba,zeng2025reverse} have emerged in the field of medical image segmentation. SegMamba~\cite{xing2024segmamba} takes the whole volumetric data as a linear sequence and explores the forward, reverse, and inter-slice features interactions. However, such linearization inherently disrupts the intrinsic 3D spatial topology. For pathological liver segmentation, where cirrhotic structures exhibit highly irregular and anisotropic morphologies, discarding the spatial context can lead to suboptimal boundary definition and fragmented results. In this paper, we propose SRMA-Mamba to model the spatial relationships within the complex anatomical structures of MRI volumes for efficient volumetric pathological liver segmentation. The main contributions of the work are as follows: 

\begin{itemize}
\item   
We present a novel spatial reverse mamba attention network, SRMA-Mamba, for efficient volumetric pathological liver segmentation in MRI volumes.

\item 
We propose a novel Spatial Anatomy-Based Mamba (SABMamba) module to perform scans within
multiple anatomical perspectives, including sagittal, coronal, and axial planes. Additionally, we introduce the Spatial Reverse Mamba Attention (SRMA) module, which progressively refines the segmentation map using hierarchical features and a coarse segmentation map.


\item 
We introduce the Anatomy-Based Selective Scan (ABSS) module to model 3D visual feature representations for pathological liver tissues, enabling direct processing of 3D volumetric data with low computational complexity. 

\item  
We conduct extensive experiments on the CirrMRI600+ T1W and T2W datasets and demonstrate that SRMA-Mamba achieves state-of-the-art performance in 3D pathological liver segmentation, surpassing existing volumetric segmentation methods.

\end{itemize}

\section{Methodology}
\subsection{Overall Architecture of SRMA-Mamba}

\begin{figure*}[!t]
    \centering
    \includegraphics[width = \textwidth]{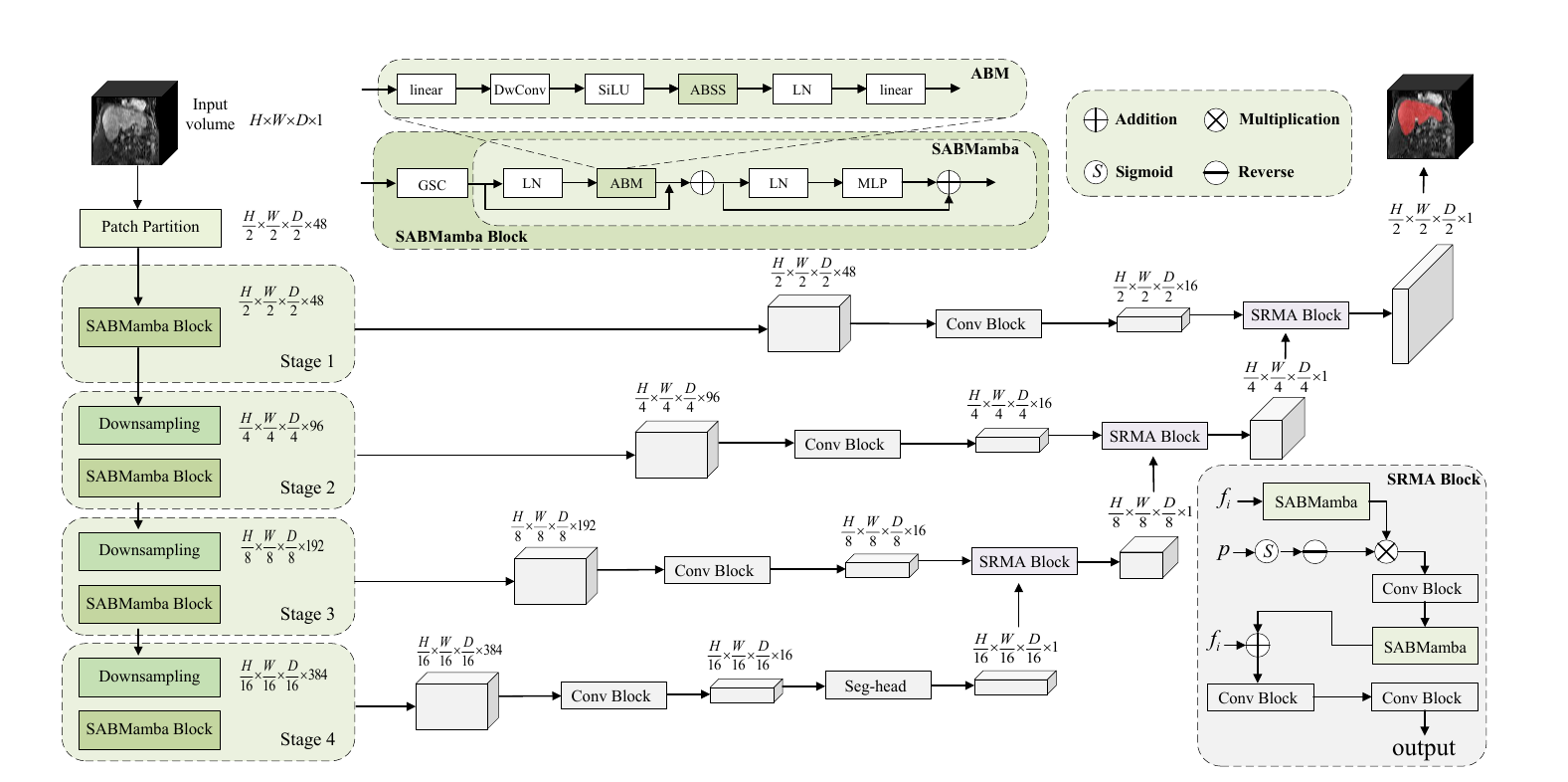}

\caption{Overview of the proposed \textit{SRMA-Mamba} architecture. }


    \label{fig:SRMA-Mamba}
\end{figure*}

SRMA-Mamba is an innovative Mamba-based network specifically designed for volumetric segmentation
of pathological liver in MRI volumes. The network leverages the Mamba-based encoder for hierarchical feature extraction and integrates a spatial reverse mamba attention module to progressively refine the segmentation map with features from each encoding stage.
As illustrated in Figure~\ref{fig:SRMA-Mamba}, the proposed network employs an encoder-decoder architecture.  
The encoder extracts different levels of features from the input cirrhotic MRI volume and the decoder incrementally refines the segmentation maps, utilizing the coarse segmentation map alongside the features obtained from the encoding stages.



\subsection{Spatial Anatomy-Based Mamba (SABMamba)}


In clinical practice, radiologists typically review MRI volumes from multiple anatomical planes to develop a comprehensive spatial awareness of liver pathology. Inspired by this, we introduce the Spatial Anatomy-Based Mamba module (SABMamba), designed to model complex anatomical relationships between spatial voxels by integrating information from three orthogonal planes. 

As shown in Figure~\ref{fig:SRMA-Mamba}, SABMamba consists of consecutive residual branches. The first branch includes a Layer Normalization followed by an ABM block, while the second branch contains a Layer Normalization and a multi-layer perceptron layer. This process can be formulated as follows:

    \begin{equation} \label{SABM1}
    x_1 = x + ABM(LN(x)) ,
    \end{equation}
    \begin{equation} \label{SABM2}
	x_2 = x_1 + MLP(LN(x_1)) ,
    \end{equation}
where MLP stands for a multi-layer perceptron, and LN refers to Layer Normalization.

In the ABM block, the input features first pass through a linear layer and a depthwise separable convolution. The resulting features are then passed through the SiLU activation function before being fed into an ABSS module. Finally, the features undergo Layer Normalization, and a linear layer is applied to produce the final output of the ABM block. 

\subsection{Anatomy-Based Selective Scan (ABSS)}

\begin{figure*}[!t]
    \centering
    \includegraphics[width = \textwidth]{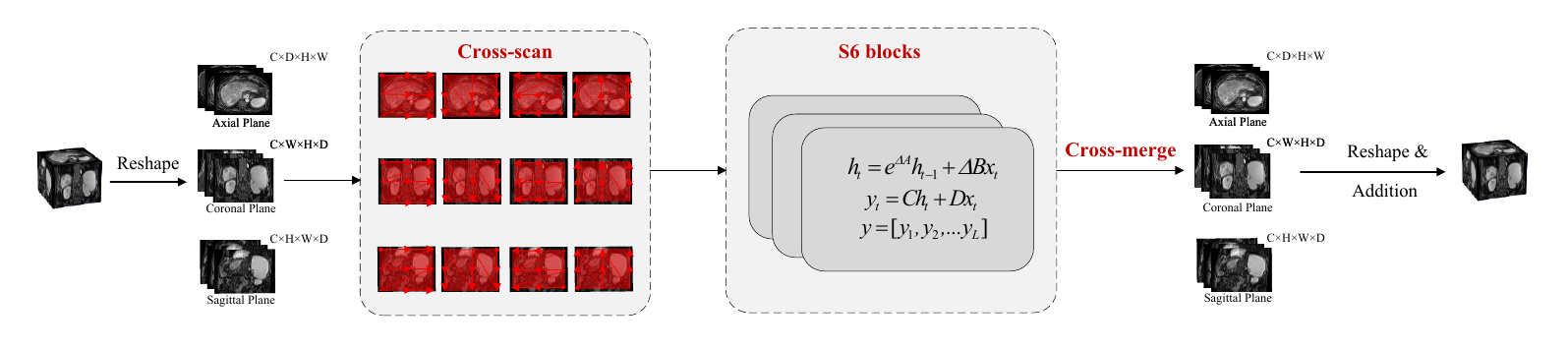}

\caption{Details of the \textit{Anatomy-Based Selective Scan (ABSS)} module}
    \label{fig:ABSS}
\end{figure*}

The overall structure of Anatomy-Based Selective Scan (ABSS) is illustrated in Figure~\ref{fig:ABSS}. ABSS is employed to learn the spatial feature representation of cirrhotic liver tissue. It comprises three key components: scan expanding operation, scan merging operation, and the S6 module~\cite{gu2023mamba}. The selective mechanism within the S6 module effectively filters out irrelevant information while preserving relevant features. 

Specifically, the input MRI volume is initially rearranged into three anatomical planes. 
\begin{equation}  
	\left\{
	\begin{array}{rcl}
f_h &= \text{Reshape}(x) \in \mathbb{R}^{C \times H \times L_1}, \\
f_w &= \text{Reshape}(x) \in \mathbb{R}^{C \times W \times L_2}, \\
f_d &= \text{Reshape}(x) \in \mathbb{R}^{C \times D \times L_3}.
	\end{array}
	\right.
\end{equation}
where \( L_1 = W \times D, L_2 = H \times D, L_3 = H \times W \).

The scan expanding operation then unfolds the data from these planes into sequences along four distinct directions. Next, the S6 module selectively processes these sequences to extract features from the anatomical planes.
After that, the scan merging operation integrates these features and restores them to their original resolution. This process can be summarized as follows:
\begin{equation} \label{SS3D}
Y = \text{CrossMerge}\left( \text{S6}\left( \text{CrossScan}(f) \right) \right)
\end{equation}
where \( f \in \{f_h, f_w, f_d\} \), and \(\text{CrossScan}\) denotes the scan expanding operation, which rearranges the input features along multiple plane-wise scanning paths, 
and \(\text{CrossMerge}\) denotes the scan merging operation, which integrates the processed sequences from the S6 blocks into the final output feature map~\cite{liu2024vmamba}.


\subsection{Spatial Reverse Mamba Attention (SRMA)}

As depicted in Figure~\ref{fig:SRMA-Mamba}, the architecture of the Spatial Reverse Mamba Attention (SRMA) consists of SABMamba and convolutional blocks. SRMA takes the coarse segmentation map and feature map from the encoding stage as input, progressively refining edge details and enhancing feature representation. Specifically, the feature map from the current level and the segmentation map from the previous level are first fed into the SRMA module. Then the segmentation map undergoes a Sigmoid function and a reverse attention operation to generate the reverse attention map weights \( A_i \), which can be formulated as:
\begin{equation} \label{reverse}
A_i = 1 - \mathrm{Sigmoid}(S_{i+1}),
\end{equation}
where \( i \in \{1, 2, 3\} \), $S_{i+1}$ denotes the coarse segmentation map. Subsequently, element-wise multiplication is applied between the attention weights and the enhanced feature map. The resulting features $F$ are further processed by a SABMamba module, which is summarized as follows:
\begin{equation} \label{reverse2}
F = \delta(Conv(A_i \cdot \delta(f_i)))
\end{equation}
where $\delta(\cdot)$ denotes a SABMamba module. Finally, a residual connection is applied, followed by two convolutional layers to generate the output.

\section{Experiments}
\label{sec:experiments}

\subsection{Dataset}
CirrMRI600+~\cite{jha2025large} is a single-center, multi-vendor, multi-sequence dataset designed to enhance cirrhotic liver research in MRI volumes. The dataset includes 628 abdominal MRI scans, consisting of 310 T1-weighted (T1W) images and 318 T2-weighted (T2W) images. For the T1W dataset, 248 images were split for training, 31 for validation, and 31 for testing. Similarly, for the T2W dataset, 256 images were allocated for training, 31 for validation, and 31 for testing. 

\subsection{Implementation details}
We implemented our model using the PyTorch framework on a single Tesla V100 GPU with 32 GB of memory. During the training phase, each volume is randomly cropped to a size of 224×224×64. The Adam optimizer is employed to optimize the model parameters with the initial learning rate of 1e-4. In our experiments, all models are trained for 500 epochs with an
early stopping patience of 50. The batch size was set to
2. The data augmentation techniques, including spatial cropping, flipping, and intensity scaling, were employed to mitigate overfitting. We use Dice loss and cross-entropy loss to supervise the network training. We use the mean Dice similarity coefficient (mDSC), mean intersection over union (mIoU), recall, precision, Hausdorff distance (HD95), and average symmetric surface distance (ASSD) to evaluate the performance of our network.


\subsection{Quantitative Comparison}
To evaluate both the efficiency and effectiveness of the proposed SRMA-Mamba, we conducted a comparative analysis against state-of-the-art volumetric segmentation methods, including those based on CNNs (SegResNet~\cite{myronenko20183d}, UX-Net~\cite{lee20223d}, MedNeXt~\cite{roy2023mednext}), Transformers (SwinUNETR~\cite{hatamizadeh2021swin}, SwinUNETRv2~\cite{he2023swinunetr}), and Mamba (SegMamba~\cite{xing2024segmamba}). As demonstrated in Table~\ref{tab:segmentationT1W} and Table~\ref{tab:segmentationT2W}, we evaluated the performance of SRMA-Mamba on the CirrMRI600+ T1W and T2W datasets, respectively. To ensure a fair comparison, all methods were trained and tested using the same codebase, dataset partitioning, and hyperparameter tuning procedures. On the T1W dataset, SRMA-Mamba exhibited outstanding performance, surpassing all other leading methods across key metrics, including mDSC, HD95, and ASSD. Specifically, our model achieves mDSC and mIoU scores of 92.95\% and 87.59\% on the T1W dataset, outperforming SegMamba by 1.15\% and 1.89\%, respectively. SRMA-Mamba also achieves impressive results with HD95 and ASSD values of 17.52 and 3.05, respectively. On the T2W dataset, SRMA-Mamba outperformed SegMamba with an impressive mDSC score of 86.25\% and an IoU score of 78.24\%, demonstrating its superior ability to delineate pathological liver structures. Additionally, compared to SegMamba, SRMA-Mamba demonstrates lower computational complexity, as shown in Table~\ref{tab:complexity}, while delivering superior segmentation results.

\begin{table*}[t!]
\centering
\caption{Model performance on the CirrMRI600+ T1W Liver
Cirrhosis MRI dataset~\cite{jha2025large}.}  
 \begin{tabular} {l|c|c|c|c|c|c}
\toprule

\textbf{{Method}}    &\textbf{mIoU} & \textbf{mDSC}  &\textbf{Recall} & \textbf{Precision} & \textbf{{HD95}} & \textbf{{ASSD}}
\\ 
\hline

SegResNet  &81.14 &88.65 &94.66 & 84.85 & 34.63	&7.55  \\

UX-Net  &79.12 &87.43 &91.21 & 84.77 &47.52 &7.19 \\


MedNeXt &82.49  &89.97 &93.62 &87.53 &29.41 &4.75 \\


SwinUNETR   &80.61 &88.71 &92.73 & 86.13 &36.79 &4.66 \\

SwinUNETRv2
 &79.68	&87.30  &89.99 &85.87 &33.27 &6.07 \\

SegMamba  &85.70 &91.80 &95.56 & 88.98 &26.97 &4.69     \\


\textbf{SRMA-Mamba} &\textbf{87.59} &\textbf{92.95} &\textbf{97.13} &\textbf{89.77} &\textbf{17.52} &\textbf{3.05}  \\

\bottomrule
\end{tabular}
\label{tab:segmentationT1W}
\end{table*}

\begin{table*}[t!]
\centering
\caption{Model performance on the CirrMRI600+ T2W Liver
Cirrhosis MRI dataset~\cite{jha2025large}.}  
 \begin{tabular} {l|c|c|c|c|c|c}
\toprule

\textbf{{Method}}    &\textbf{mIoU} & \textbf{mDSC}  &\textbf{Recall} & \textbf{Precision} & \textbf{{HD95}} &
\textbf{{ASSD}} \\ 
\hline


SegResNet   &76.74 &84.86 &86.55 &86.93 &28.47 &4.60 \\

UX-Net 
&72.35 &82.01 &84.33 &84.80 &43.87 &5.86  \\

MedNeXt &69.03  &79.87 &83.32 &81.28 &44.83 &6.33  \\


SwinUNETR  &74.90 &83.91 &87.01 &84.83 &34.61 &4.84  \\

SwinUNETRv2 &75.21  &84.89 &85.51 &86.96 &29.37 &\textbf{4.26} 	\\

SegMamba   &76.99 &85.36 &87.41 & 87.00 &30.75 &4.52  \\

\textbf{SRMA-Mamba}   &\textbf{78.24} &\textbf{86.25} &\textbf{87.71} &\textbf{87.52} &\textbf{26.04} &4.57  \\

\bottomrule
\end{tabular}
\label{tab:segmentationT2W}
\end{table*}



\subsection{Qualitative Comparison}
\begin{figure*} [!t]
    \centering
    \includegraphics[width=0.9\textwidth]{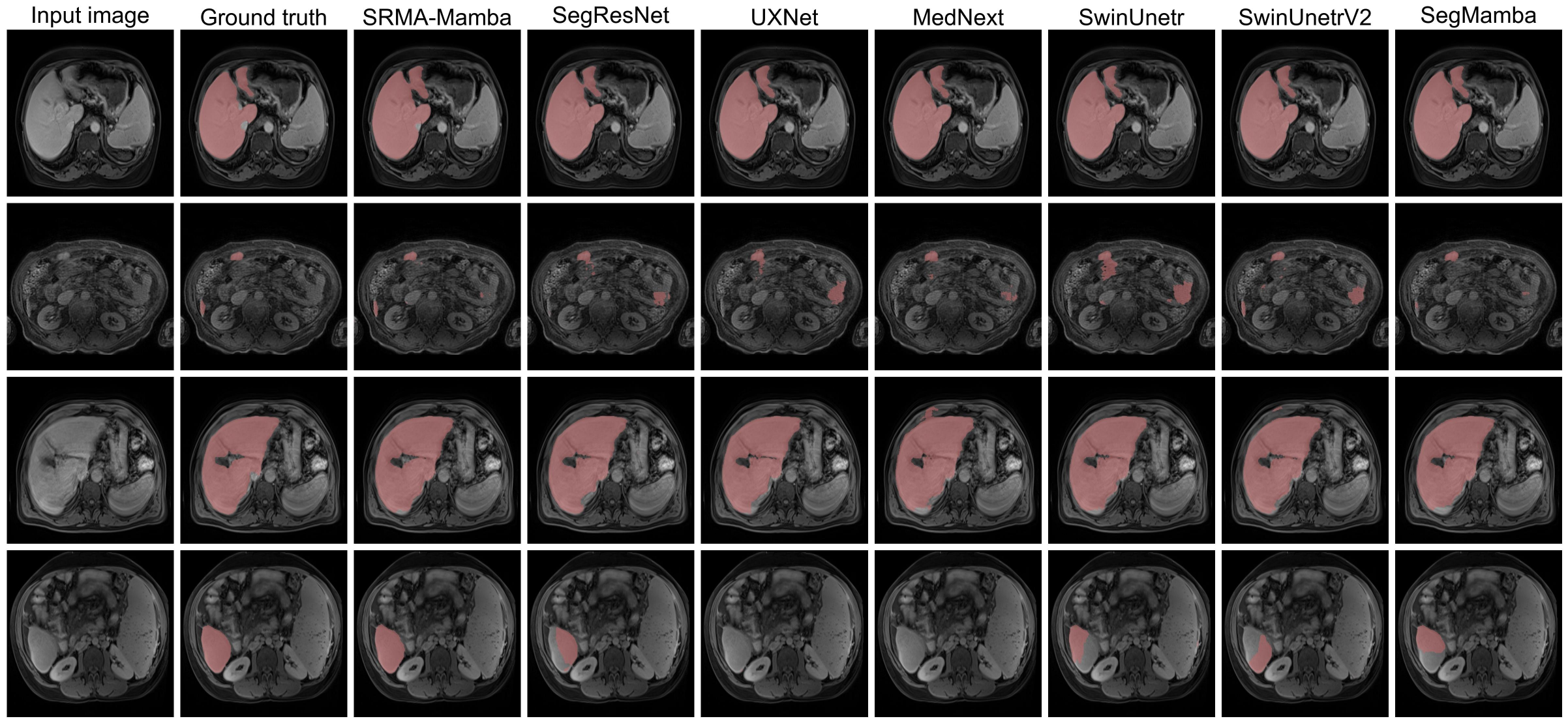}
    \caption{Qualitative results of different methods on the CirrMRI600+ T1W dataset.}
    \label{fig:visualizationT1W}
\end{figure*}

\begin{figure*} [!t]
    \centering
    \includegraphics[width=0.9\textwidth]{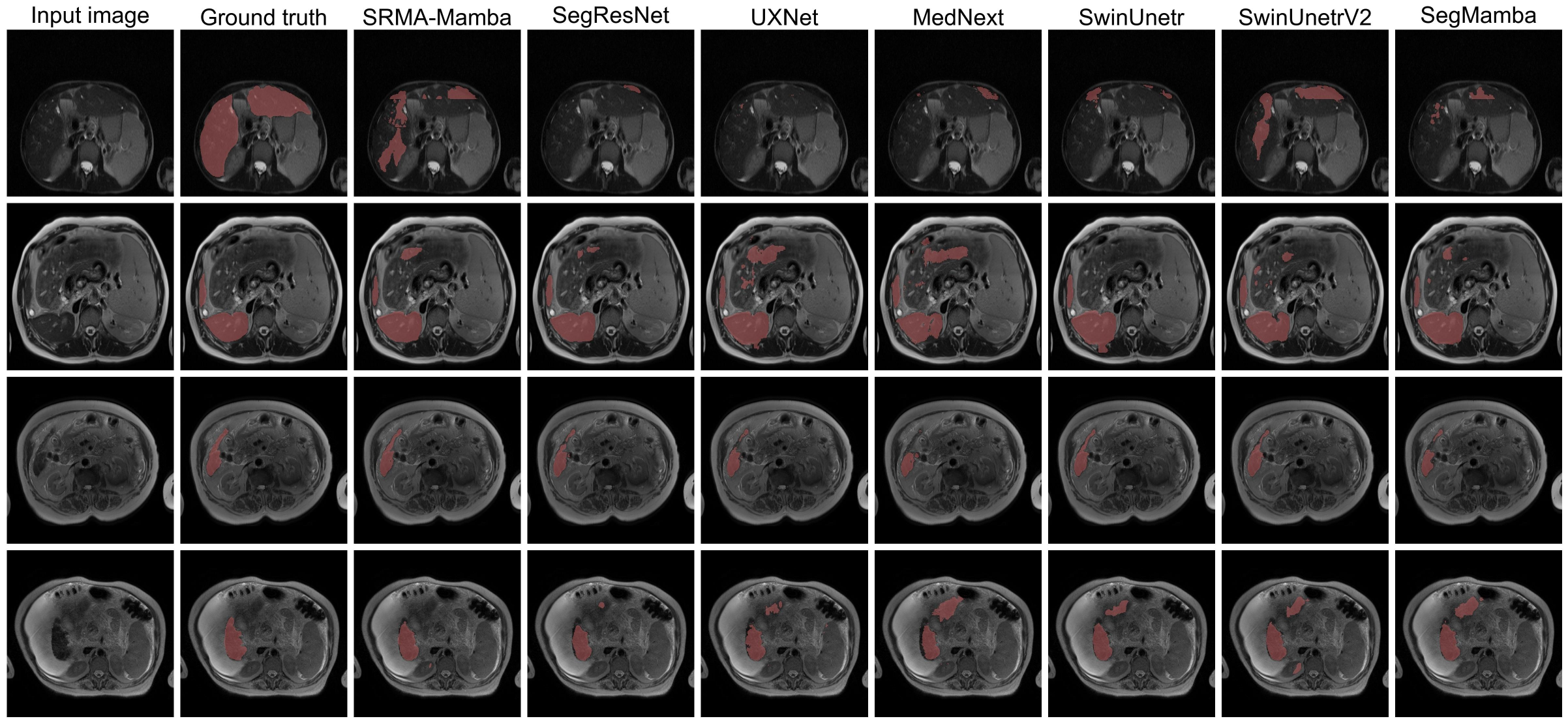}
    \caption{Qualitative results of different methods on the CirrMRI600+ T2W dataset.}
    \label{fig:visualizationT2W}
\end{figure*}

Figure~\ref{fig:visualizationT1W} and Figure~\ref{fig:visualizationT2W} present a qualitative comparison of state-of-the-art methods on the CirrMRI600+ T1W and T2W datasets, highlighting the superior capability of SRMA-Mamba in delineating pathological liver structures. The proposed SABMamba effectively models spatial voxel dependencies by integrating feature information across multiple anatomical planes. Additionally, SRMA-Mamba achieves impressive segmentation performance by optimizing boundary details at multiple decoding stages. On the CirrMRI600+ T1W dataset, SRMA-Mamba demonstrates superior performance in segmenting both large and small pathological regions. Despite the challenging cases in the CirrMRI600+ T2W dataset, which occasionally result in false positives and missed detections of cirrhotic tissues, SRMA-Mamba consistently outperforms most state-of-the-art models, showcasing exceptional segmentation capabilities.

\begin{table}[t!]
    \centering
    \renewcommand{\arraystretch}{1.1}
    \setlength\tabcolsep{5pt}
    \caption{Computational complexity of various methods.}
    \label{tab:complexity}
    \begin{tabular}{l|cc}
        \toprule    
        \textbf{Method} 
        & \textbf{Parameters} 
        & \textbf{GFLOPs}   \\
        
        \hline
        
        SegResNet
        & 1.17M
        & 55.95      \\
        
        UX-Net
        & 53.00M
        & 2292.44 \\
        
        MedNeXt 
        & 5.54M
        & 210.91   \\
        
        SwinUNETR
        & 15.51M
        & 303.33  \\
                
        SwinUNETRv2
        & 18.15M
        & 326.79  \\
        
        SegMamba
        & 64.24M
        & 2379.03  \\
        
        \textbf{SRMA-Mamba}
        & 17.22M
        & 149.14 \\
        \bottomrule
    \end{tabular}
\end{table}

\begin{table}[t!]
\centering
\caption{Ablation study on the CirrMRI600+ T1W dataset~\cite{jha2025large}. \ding{51} denotes the module is included,
\ding{55} denotes the module is removed, corresponding to the “w/o” notation in the text.}  
\setlength{\tabcolsep}{8pt}
\renewcommand{\arraystretch}{1.1}

\begin{tabular}{c|c@{\hspace{3pt}}c@{\hspace{4pt}}|cc}
\toprule

{\textbf{Exp.}} &
{\textbf{SABMamba}} &
{\textbf{SRMA}} &
\textbf{mIoU} &
\textbf{mDSC} \\
\hline

\#1  &\ding{55} &\ding{55} &46.66 &61.65    \\ 

\#2  &\ding{55} &\ding{51}  &68.61 &80.62    \\ 



\#3  &\ding{51} &\ding{55}  &83.82 &90.63    \\ 

\#4 &\ding{51} &\ding{51} &\textbf{87.59} &\textbf{92.95}   \\ 

\bottomrule
\end{tabular}
\label{tab:ablationT1W}
\end{table}

\begin{table*}[t!]
\centering
\caption{
Performance comparison of different anatomical plane configurations on the CirrMRI600+ T1W dataset~\cite{jha2025large}.
}
\setlength{\tabcolsep}{8pt}
\renewcommand{\arraystretch}{1.1}
\begin{tabular}{l|ccc}
\toprule
\textbf{Anatomical planes}
& \textbf{mIoU}
& \textbf{mDSC}
& \textbf{HD95} \\
\hline

Axial     &86.61 &92.48 &18.92 \\
Sagittal  &86.69 &92.50 &19.56 \\
Coronal   &82.78 &90.22 &27.38 \\

\textbf{Three Planes (Ours)}
&\textbf{87.59}
&\textbf{92.95}
&\textbf{17.52} \\

\hline
\end{tabular}
\label{tab:ablationOfAnatomicalPlanes}
\end{table*}

\subsection{Ablation Study}

To evaluate the effectiveness of the proposed modules, we perform comprehensive ablation experiments on the CirrMRI600+ T1W dataset. Table~\ref{tab:ablationT1W} presents the ablation results of SABMamba and SRMA. “w/o SABMamba” replaces the SABMamba encoder with the UNETR encoder, while “w/o SRMA” replaces the SRMA decoder with a standard 3D convolutional decoder. Replacing the SABMamba encoder with the UNETR encoder decreases the mDSC from 92.95\% to 80.62\%, demonstrating the effectiveness of SABMamba in modeling spatial anatomical information. In addition, removing SRMA reduces the mDSC from 92.95\% to 90.63\%, indicating that SRMA provides complementary improvements by enhancing feature aggregation and boundary delineation. 
The results in Table~\ref{tab:ablationOfAnatomicalPlanes} further validate the effectiveness of the proposed ABSS module. By integrating information from axial, sagittal, and coronal planes, the three-plane configuration consistently outperforms all single-plane variants, demonstrating the benefit of multi-plane anatomical representation learning.

\section{Conclusion}
This paper presents a novel spatial reverse mamba attention network, named SRMA-Mamba, for volumetric segmentation of pathological liver in MRI volumes. We introduce the SABMamba module, which performs planar scanning across the three anatomical planes and combines feature information from these planes to construct a spatial representation of liver cirrhosis regions. Additionally, SRMA-Mamba progressively refines boundary precision through multiple Spatial Reverse Mamba Attention modules, thereby enhancing the overall
segmentation performance. We also introduce an Anatomy-Based Selective Scan module that directly processes voxel data to generate 3D feature representations, preserving spatial anatomical information. Extensive experiments demonstrate that SRMA-Mamba achieves state-of-the-art performance on both T1W and T2W modalities of the liver cirrhosis benchmark dataset, surpassing existing methods and highlighting its effectiveness and stability in the volumetric segmentation of pathological liver.

\begin{credits}
\subsubsection{Acknowledgments}
D. Jha is supported in part by the U.S. Department of Education (P116Z240151 to the University of South Dakota and the South Dakota School of Mines \& Technology). The views expressed are those of the author(s) and do not necessarily represent the official views of the U.S. Department of Education.

\subsubsection{Disclosure of Interests}
The authors have no competing interests to declare.
\end{credits}

%
%
%
%
\bibliographystyle{splncs04}
\bibliography{reference}





\end{document}